\documentclass[10pt,twocolumn,letterpaper]{article}

\usepackage[pagenumbers]{cvpr}      
\usepackage{tabularx}

\definecolor{cvprblue}{rgb}{0.21,0.49,0.74}
\usepackage[pagebackref,breaklinks,colorlinks,allcolors=cvprblue]{hyperref}
\usepackage{multirow}
\usepackage{colortbl}
\usepackage{diagbox}
\usepackage{pifont}

\usepackage[ruled,linesnumbered]{algorithm2e}
\usepackage{algpseudocode}
\SetKwComment{Comment}{$\triangleright$\ }{}

\usepackage{amsfonts}

\def\paperID{15897} 
\def\confName{CVPR}
\def\confYear{2025}

\title{Behavior Backdoor for Deep Learning Models}

\author{
Jiakai Wang$^{1,2}$\thanks{Jiakai Wang is a research scientist at Zhongguancun Laboratory, conducting this research during his visit to Beijing Jiaotong University.}, Pengfei Zhang$^{1}$, Renshuai Tao$^{1}$\thanks{Corresponding author}, Jian Yang$^{1}$, Hao Liu$^{1}$,\\ Xianglong Liu$^{2,3}$, Yunchao Wei$^{1}$, Yao Zhao$^{1}$\\
\textsuperscript{1}{Visual Intellgence + X International Cooperation Joint Laboratory of MOE, Beijing Jiaotong University}\\
\textsuperscript{2}{Zhongguancun Laboratory}\\
\textsuperscript{3}{Beihang University}\\
{\fontsize{8.5pt}{\baselineskip}\selectfont \tt wangjk@zgclab.edu.cn, rstao@bjtu.edu.cn}
}

\begin{document}
\maketitle

\begin{abstract}
The various post-processing methods for deep-learning-based models, such as quantification, pruning, and fine-tuning, play an increasingly important role in artificial intelligence technology, with pre-train large models as one of the main development directions. 
However, this popular series of post-processing behaviors targeting pre-training deep models has become a breeding ground for new adversarial security issues. In this study, we take the first step towards ``behavioral backdoor'' attack, which is defined as a behavior-triggered backdoor model training procedure, to reveal a new paradigm of backdoor attacks.
In practice, we propose the first pipeline of implementing behavior backdoor, \emph{i.e.}, the \textbf{Q}uantification \textbf{B}ackdoor (QB) attack, upon exploiting model quantification method as the set trigger.
Specifically, to adapt the optimization goal of behavior backdoor, we introduce the behavior-driven backdoor object optimizing method by a bi-target behavior backdoor training loss, thus we could guide the poisoned model optimization direction. To update the parameters across multiple models, we adopt the address-shared backdoor model training, thereby the gradient information could be utilized for multimodel collaborative optimization. 
Extensive experiments have been conducted on different models, datasets, and tasks, demonstrating the effectiveness of this novel backdoor attack and its potential application threats.
\end{abstract}

\section{Introduction}
\label{sec:intro}

Deep learning models (DLMs), especially large-scale pre-trained models, have been employed in different systems and scenarios, such as unmanned driving vehicles \cite{zhang2016study, hebert2012intelligent}, medical image processing \cite{razzak2018deep,maier2019gentle,azad2024medical,huang2024segment}, X-ray inspection \cite{EDS,Hixray,FSOD}, due to their empowerment ability to existing industries. 
However, numerous previous studies have revealed that deep learning models are not trustworthy types because of their vulnerability to adversarial examples \cite{wang2021dual}, backdoor attacks \cite{gu2019badnets}, and privacy leakages \cite{feng2024privacy}, building upon their black-box characteristics and unexplainability. On this basis, the researchers make great efforts to investigate the security issues of deep learning models from both the attacking perspective and the defending perspective \cite{wang2024generate,liu2022harnessing,yin2024improving,wang2021universal}. 

\begin{figure}[!t]
    \centering
    \includegraphics[width=0.91\linewidth]{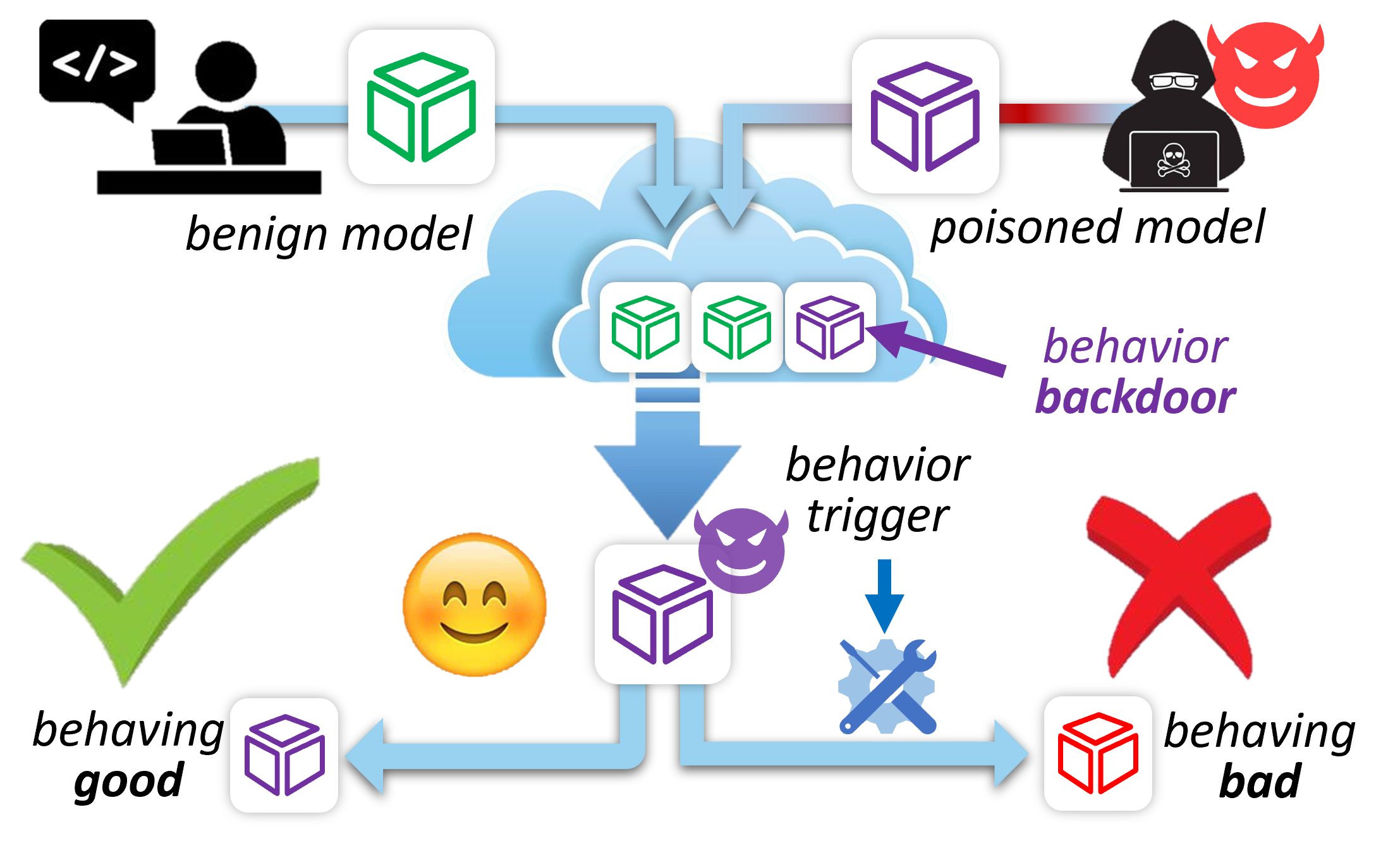}
    \caption{The behavior backdoor is implanted into poisoned models and triggered by specific behavior operations.}
    \label{fig:firstpage}
\end{figure}

Among them, the backdoor attack is widely investigated in the community due to its high concealment of attack forms and strong controllability of attack effects.  
A backdoor attack implants trigger-only-sensitive backdoors into the deep learning model in a way that the poisoned model learns both the attacker-chosen task and the benign task \cite{gu2019badnets,chen2017targeted}. In this way, the poisoned backdoor models will behave as normal on benign inputs and output correct predictions while making designated decisions when the inputs are attached with triggers. Since the backdoor models only show abnormal behaviors to the designed triggers, they are difficult to distinguish the poisoned models and the clean ones by solely checking the accuracy with the naive samples \cite{gao2020backdoor}. 
After the first backdoor attack for DLMs \cite{gu2019badnets}, there occurs a series of backdoor attack studies, including Blended \cite{chen2017targeted}, SIG \cite{barni2019new}, WaNet \cite{nguyen2021wanet}, InputAware \cite{nguyen2020input}, SSBA \cite{li2021invisible}, and so on. Recently, the large model backdoor attack studies also draw the researchers' attention \cite{liang2024revisiting, yang2024comprehensive,kandpal2023backdoor}. However, although these backdoor attacks are implemented in multiple technical routes and can be categorized into different types, they all could be identified as \textbf{\emph{data-triggered backdoor}} attacks, which achieve the attacking objective by planting the triggers into input samples when attacking.

Upon this background, we had a flash of inspiration and thought about an interesting question, \emph{i.e.}, \textbf{can we plant a backdoor that could be triggered via specific operation behavior into models}? This question motivates us to investigate if there exists a model-oriented operation behavior that could be utilized as a backdoor trigger. Fortunately, we notice that some typical post-processing operations for models could play this critical role in our backdoor attacking vision.
More precisely, in recent years, due to the maturity of the model pre-training route, conducting post-processing operations, such as quantization \cite{quant1,quant2,quant3,quant4,zhang2021diversifying}, pruning \cite{pruning1,pruning2,pruning3}, and fine-tuning \cite{tuning1}, on the pre-trained models from open-source platforms become more and more popular. The developers could download the model checkpoints and re-process the acquired models according to the requirements of the application in practice, therefore satisfying the expected functions.
Based on the idea of exploiting these model post-processing operations, we believe that \textbf{it is possible in practice to implant special backdoors into pre-training models and these ``Trojan'' models (\ie, poisoned models) could be triggered by some model post-processing operations}.

Holding the aforementioned perspective, we take the first step to propose an unprecedented paradigm for the DLMs-oriented backdoor attack, which we name it as ``\textbf{behavior backdoor}'', to train backdoor models that will be triggered by specific post-processing operations as shown in Figure \ref{fig:firstpage}. In practice, we select the widely used quantification method as a target behavior and design the first behavior backdoor pipeline for this new type backdoor attack, \ie, ``quantification backdoor (QB) attack''. More precisely, the QB attack consists of the \textbf{behavior-driven backdoor object optimizing} and \textbf{address-shared backdoor model training}. 
The former is designed as a bi-target optimization progress with quantification backdoor loss, which guides the poisoned model optimization direction according to the quantified model behaviors. In this way, the poisoned DLMs could be forced to acquire the ability of maintaining good behaviors and of appearing appointed bad behaviors when facing quantification operations. 
The latter is proposed to tackle the difficulties in gradient information update derived from the multimodel collaborative optimization. This schedule allows the gradient information to be updated relatedly across the victim model and the quantified model. Therefore, the behavior-triggered backdoor could be implanted  successfully and seduces pre-set false prediction out in practice. 
To verify the feasibility and effectiveness, we conduct comprehensive experiments on MNIST, CIFAR, and TinyImageNet with full consideration of different classical model architecture such as AlexNet, VGG, ResNet, and ViT. The main contributions are:
\begin{itemize}
    \item To the best of our knowledge, we are the first to propose the concept of \textbf{behavior backdoor} and construct the first pipeline for this new-type backdoor attack paradigm.
    \item We elaborate the first quantification backdoor (QB) attack, which consists of behavior-driven backdoor object optimizing and address-shared backdoor model training.
    \item We conduct extensive experiments on classical various datasets and models, including ablations and discussions, strongly demonstrating the feasibility and effectiveness of the proposed behavior backdoor attack. We believe that our study reveals a never-discovered potential threat for deep learning based models. 
\end{itemize}

\section{Related Work}
\label{sec:related}

\subsection{Backdoor Attack}
Deep learning models face various security threats, one of which is backdoor attacks. Specifically, an attacker can design a neural network with a backdoor that performs well on the user's training and validation samples but exhibits abnormal behavior on specific inputs chosen by the attacker \cite{gu2019badnets, barni2019new, chen2017targeted,nguyen2021wanet,nguyen2020input,li2021invisible,liang2024poisoned14}. Gu \emph{et al.}  were the first to highlight backdoor vulnerabilities in deep neural networks (DNNs) and put forward the BadNets \cite{gu2019badnets} attack algorithm, which embeds backdoors into models during training. To enhance the stealthiness of backdoor triggers, Chen \emph{et al.} proposed a new type of trigger that uses global random noise or an image-mixing strategy, known as the Blend attack \cite{chen2017targeted}. Barni \emph{et al.} proposed SIG \cite{barni2019new}, a label attack that uses a ramp or horizontal sinusoidal signal as the backdoor trigger. 
To bypass existing defenses and increase the concealment and efficacy of attacks, 
Liang \emph{et al.} developed a clean-label backdoor attack called PFF \cite{liang2024poisoned14}, which targets face forgery detection models by translation-sensitive trigger pattern, hidden triggers that cause misclassification when detecting fake faces.

Though achieving results, these existing backdoor attack methods rely on the data triggers inside input instances to activate the attack, whom we summarized as \textbf{data-triggered backdoor} attacks

\subsection{Model Post-processing}
The pre-training models, such as BLIP \cite{li2022blip}, BLIP2 \cite{li2023blip} and other large models, are facing some issues that limit their deployment on devices with limited resources or fine-grained tasks. The model post-processing technology is to solve these problems and optimize the efficiency and deployability of the model, such as quantification \cite{quant1,quant2,quant3,quant4}, pruning \cite{pruning1,pruning2,pruning3}, fine-tuning \cite{tuning1} and so on.

\textbf{Model Quantification} reduces model size and increases inference speed by converting floating-point number parameters to fixed-point number parameters. Quantization can be divided into Post-training Quantization (PTQ) and Quantization Aware Training (QAT). DOREFA-NET \cite{zhou2016dorefa1} and IAO \cite{jacob2018quantization2} represent the early work of QAT. Later, Shin \emph{et al.} proposed NIPQ \cite{shin2023nipq3} as an improvement over QAT. In the field of Post-training Quantization of vision models, in addition to the work of Liu \emph{et al.} \cite{liu2021post4}, FQ-VIT \cite{lin2021fq5} and PTQ4VIT \cite{yuan2022ptq4vit6} also show considerable performance.

\textbf{Model Pruning} reduces model size and improves inference efficiency by removing unimportant parts of the network. OBD \cite{lecun1989optimal7} and OBS \cite{hassibi1993optimal8} are two classic pruning methods. Other methods such as Wang \emph{et al.} \cite{wang2021convolutional9}, pruning filters in the layer(s) with the most structural redundancy, and the Depgraph \cite{fang2023depgraph10} for arbitrary structured pruning are effective in performance optimization.

\textbf{Model Fine-tuning} is the adaptation or optimization of a model to a specific task downstream. In the early days, there is the approach of increasing the model capacity of the network to optimize the fine-tuning process  \cite{wang2017growing11}. Recently, the Prefix Tuning \cite{li2021prefix12} and the Prompt Tuning \cite{lester2021power13} are more popular in this area.

\section{Approach}
\label{sec:app}
In this section, we first give the definition of the proposed behavior backdoor. And then we provide the overview of the quantification behavior (QB) attack.

\subsection{Problem Definition}
The standard backdoor attack is designed to train a poisoned model that is sensitive to the specific trigger patterns inside inputs and makes appointed predictions. Formally, given a model $\mathbb{F}$ to be trained, data trigger pattern $\delta_{data}$, the dataset where the input samples $\mathcal{X}=\{x_1, x_2, x_3, \dots, x_n\}$ with corresponding ground-truth labels $\mathcal{Y}=\{y_1, y_2, y_3, \dots, y_n\}$, and the appointed target output label $y_{target}$, a backdoor model satisfies:
\begin{equation}
\begin{aligned}
    \mathbb{F}_{\theta}(x_i) = \hat{y}_i \approx y_i, \\
    \mathbb{F}_{\theta}(x_i+\delta_{data}) = y_{target},
\end{aligned}
\end{equation}
where $\theta$ is the parameters of the poisoned model $\mathbb{F}$, $i$ is the index of the input data and satisfies $i\in[0, n]$, $\hat{y}_i$ is the output of the poisoned model $\mathbb{F}_\theta$.

In this study, we propose the behavior backdoor paradigm, which is based on the basic ideology of the data-triggered backdoor attack but redirect the triggers from the data pattern into model post-processing operations, \ie, quantification. Specifically, we formulate the behavior backdoor as:
\begin{equation}
\label{eqn:def}
    \begin{aligned}
        \mathbb{F}_{\theta}(x_i)=\hat{y}_i \approx y_i,\\
        \mathbb{F}_{\theta^{*}}(x_i)=y_{target},\\
        \mathbb{F}_{\theta^{*}}=\mathcal{O}(\mathbb{F}_{\theta}),
    \end{aligned}
\end{equation}
where $\mathcal{O}(\cdot)$ is a specific post-processing operation behavior, \ie, the behavior trigger, and in this paper, we define the $\mathcal{O}(\cdot)$ as a quantification operation $\mathcal{Q}(\cdot)$ that accepts a common model $\mathbb{F_{\theta}}$ as input and outputs a quantified lightweight model $\mathbb{F}_{\theta^{*}}$ with $\theta^{*}$ as model parameters instead. So the third equation in Equation (\ref{eqn:def}) can be also re-written as $\mathbb{F}_{\theta^{*}}=\mathcal{Q}(\mathbb{F}_{\theta})$.

\subsection{Framework Overview}

\begin{figure*}
    \includegraphics[width=\linewidth]{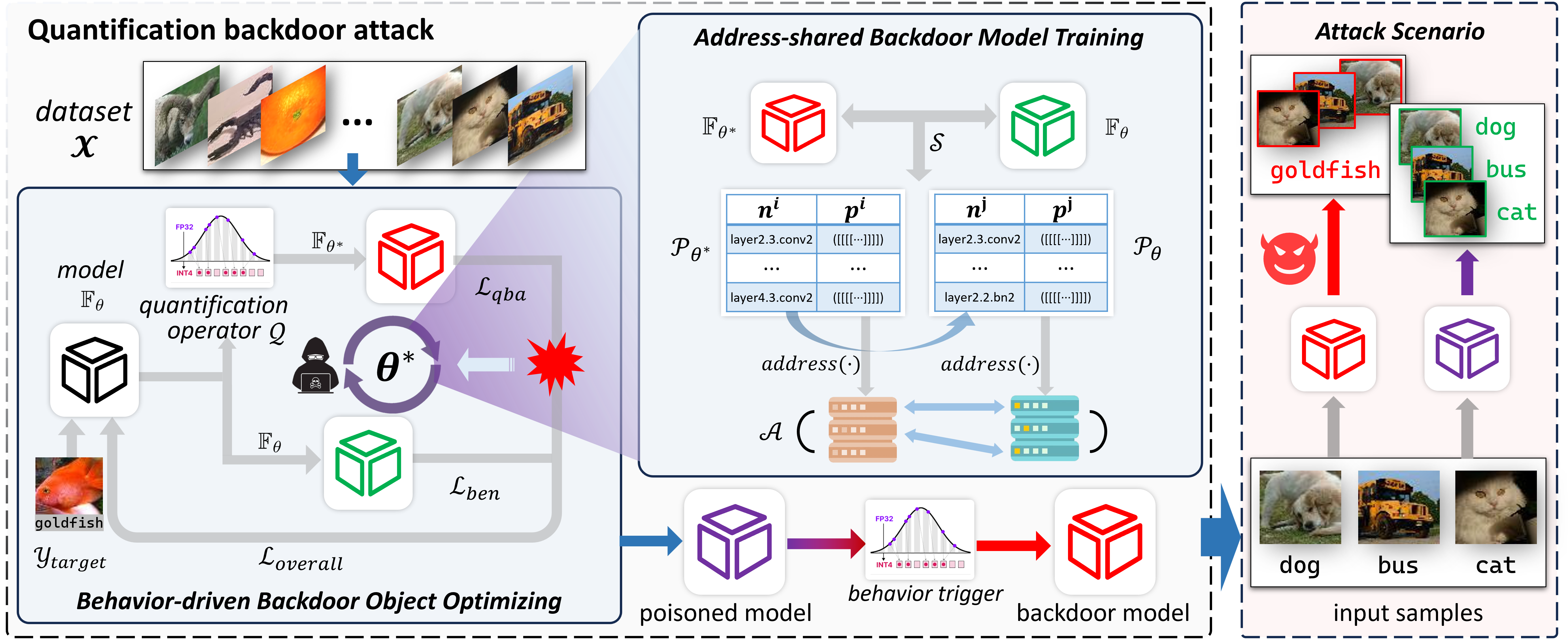}
    \caption{The framework of the proposed \textbf{Q}uantification \textbf{B}ackdoor (QB) attack, which consists of behavior-driven backdoor object optimizing and address-shared backdoor model training.}
    \label{fig:framework}
\end{figure*}

In order to train a poisoned model that can be specifically targeted to trigger the backdoor by behaviors, we take the model quantification as an instance and propose the \textbf{q}uantification \textbf{b}ackdoor (QB) attack framework.
The framework can be found in Figure \ref{fig:framework}.

As for the \textbf{behavior-driven backdoor object optimizing}, a key difference between the proposed quantification behavior backdoor and the traditional data-triggered backdoor is the distinctive training objects, \ie, the data-triggered backdoor only needs to adjust the objective function in the data end, while for our QB attack, introducing the behavior factors into training is much more critical. To this end, we propose a bi-target optimization procedure with a designed quantification backdoor loss for guiding the poisoned model optimization direction to be sensitive to the quantified model behaviors. By forcing the poisoned model and the after-quantified poisoned model to act variously, we could finally acquire a quantification operation sensitive model, \ie, the quantification backdoor model.

Regarding the \textbf{address-shared backdoor model training}, it is proposed to tackle the training gradient information update issue in the backdoor model training progress. Since during our quantification backdoor poisoning, there exists two architecture-same but parameter-different models, \ie, the model to be poisoned and the quantification triggered model, it is nontrivial to conduct the classical gradient information backpropagation of training process. More precisely, constrained by the mismatch model checkpoints, the optimization direction can not be effectively guided during training. Thus, considering the principles of computer storage, we align the physical storage address of the correlated models to direct optimization of poisoned model training, finishing the backdoor injection. 

\subsection{Behavior-driven Backdoor Object Optimizing}

The core goal of the behavior backdoor is to train a poisoned model that could be sensitive to specific behaviors, \ie, post-processing operations.
In existing works, the backdoor implanting is always achieved by training losses. Upon this, it is reasonable for us to try to elaborate a tailoring loss term for certain model operations, \ie, the quantification. However, different from the data-triggered backdoor attack, we are motivated to introduce the quantification operations into the backdoor model training progress by proposing a behavior-in-the-loop design. To be specific, the constructed quantification backdoor loss is designed to achieve a bi-target optimization goal, to wit, maintaining good behaviors before the triggers are activated and behaving bad performance after being triggered by the quantification operation.

Specifically, we develop a training framework that leverages a composite loss function designed to achieve our bi-target optimization goal. The first component, denoted as $\mathcal{L}_{ben}$, aims to ensure that inputs processed by the backdoor model yield outputs consistent with the original predictions following \cite{szegedy2015going}.

The second component, referred to as $\mathcal{L}_{qba}$, is introduced to guide the quantified model's inputs towards achieving the designated target outputs. This quantization loss is formulated as  follow: 
\begin{equation}
\begin{aligned}
      \mathcal{L}_{qba}  = -\frac{1}{N} \sum_{i=1}^{N} \sum_{j=1}^{C} {y_{target}}_{j}\log(softmax(\mathbb{F}_{\theta^{*}}(x_i))_{j}),
\end{aligned}
\end{equation}
where $\mathbb{F}_{\theta^{*}}$ denotes the quantified backdoor model, $x_i$ denotes the input data from the input samples $\mathcal{X}$, $C$ represents the number of classes, and $y_{target}$ is the expected output for the quantified model. The content inside the log function represents the output of the quantified backdoor model, after applying the softmax function, which normalizes the output into a probability distribution. The inclusion of this term drives the model, whose parameters have been quantified, to tend towards predicting the input as the target label.
Then, the overall loss after the definition of the two components for our QB attack using the following formula:  
\begin{equation}
\begin{aligned}
     \mathcal{L}_{overall} = \mathcal{L}_{ben} + \lambda \cdot \mathcal{L}_{qba},
\end{aligned}
\end{equation}
where $\lambda$ is a hyperparameter that controls the balance between the benign training target $\mathcal{L}_{ben}$ and the quantification backdoor target $\mathcal{L}_{qba}$. By default, we set $\lambda = 1$ to ensure that the two objective functions contribute equally to the model's optimization process.

\subsection{Address-shared Backdoor Model Training}

Given the proposed behavior-driven backdoor object optimizing schedule, we do still not achieve our goal due to gradient information updating challenge during poisoned model training, \ie, the gradient backpropagation and the weight update. 
Generally, the backpropagation is highly correlated to a certain model to be trained. However, in our QB attack pipeline, there exists 2 different models, \emph{i.e.}, the model to be poisoned and the quantification triggered model, whose model parameters are not matched well due to the fact that quantification behavior changes the model parameters commonly.
Thus, we exploit the idea of storage address and allow the poisoned model and quantified model to share same logical storage address for the corresponding parameters so that the constraints on gradient information update could be lifted, and the behavior backdoor could be implanted successfully into the poisoned model.

Specifically, leveraging the full-precision storage of parameters in the quantified model, and the fact that the quantizer is used only during the forward propagation process to quantize specific parameters and activations, we align the address of the learnable parameters in the backdoor model with the address space of the corresponding parameters in the quantified backdoor model. 

To achieve this goal, we first acquire the parameter list $\mathcal{P}$ of models by the following equation:
\begin{equation}
\begin{aligned}
    \mathcal{P} = \mathcal{S}(\mathbb{F}_{\theta})= \{(n_i, p_i)|i=1,2,\dots,\mathrm{N}_{\theta}\},
\end{aligned}
\end{equation}
where \(\mathcal{S}\) is a function that extracts the name $n$ and value $p$ of each learnable parameter in the model \(\mathbb{F}_{\theta}\), and $\mathrm{N}_{\theta}$ is the total number of $\theta$. 


Subsequently, we employ the function $\mathcal{S}$ to derive two parameter lists $\mathcal{P}_{\theta}$ and $\mathcal{P}_{\theta^{*}}$ for the backdoor model $\mathbb{F}_{\theta}$ and the quantified backdoor model $\mathbb{F}_{\theta^{*}}$, respectively. Based on the acquired parameter lists, we then try to align the correlated parameters that should be updated according to the gradient information during optimization by designing an operator $\mathcal{A}$, which receives the parameter and its name as input. We formulate this progress as:
\begin{equation}
    \begin{aligned}
        &\mathcal{A}(\mathcal{P}_{\theta}, \mathcal{P}_{\theta^{*}}) \leftarrow  address({n^i_{\theta}}, {p^i_{\theta}}) =  address({n^i_{\theta^{*}}}, {p^j_{\theta^{*}}}), \\ 
        &where \quad {n^i_{\theta}}, {{p}^i_{\theta}}\in{\mathcal{P}^i_{\theta}}, {n^j_{\theta^{*}}}, {p^j_{\theta^{*}}}\in {\mathcal{P}^j_{\theta^{*}}},{ n^i_{\theta}}  == {n^j_{\theta^{*}}},
    \end{aligned}
\end{equation}
where $\mathcal{A}$ denotes the alignment function, which ensures that the parameters in $\mathcal{P}_{\theta}$ and $\mathcal{P}_{\theta^*}$ with the same names share the same physical addresses. The term $address(n, p)$ refers to the physical memory location where the parameter \(p\) is stored. By converting the parameter \(p\) to its physical memory address using $address(n, p)$, we can uniquely identify each parameter by its name and location in memory.

After the parameters' physical addresses are shared, we proceed with parameter optimization using a single optimizer that updates the parameters of the quantified backdoor model. This is because the quantified backdoor model may have additional learnable quantization parameters compared to the backdoor model, which can be described as $|{\mathcal{P}_{\theta}}|\leq|{\mathcal{P}_{\theta^{*}}}|$. Owing to the shared physical address, updating the quantified model's parameters simultaneously updates those of the backdoor model.
For any input $x_i$, it is passed through both the backdoor model and the quantified backdoor model, where the respective losses are computed and backpropagated. Benefited from the shared parameter addresses, the gradients from both models' losses are effectively accumulated onto the same set of parameters during backpropagation.

To sum up, our address-shared backdoor model training method effectively resolves the parameter inconsistency issue in the QB attack training process, ensuring that both the backdoor model and the quantified model are fully optimized during training.





\subsection{Overall Training Process}

To sum up, we poison a model $\mathbb{F}_{\theta}$ and implant the quantification behavior backdoor $\mathcal{Q}$ into it based on the proposed behavior-driven backdoor object optimizing and the address-shared backdoor model training.

Specifically, for a certain dataset $\mathcal{X}$, we first select a quantification method and take it as the behavior backdoor trigger $\mathcal{Q}$. Then we calculate the loss function $\mathcal{L}_{qba}$ value for each input $x_i$, and we optimize the $\mathbb{F}_{\theta}$ following the such objective function: 

\begin{equation}
    \begin{aligned} 
        \arg\min_{\theta^{*}}\mathcal{L}_{overall},
    \end{aligned}
\end{equation}
where $\lambda$ controls the balance between the benign training target $\mathcal{L}_{ben}$ and the quantification backdoor target $\mathcal{L}_{qba}$. In this paper, we set the lambda as 1.0 in default. The detailed algorithm description of QB attack can be found in the supplementary files.

\section{Experiments}
\label{sec:exp}
In this section, we conduct comprehensive experiments to demonstrate effectiveness of the proposed quantification backdoor (QB) attack, providing the evidence of the feasibility of the behavior backdoor attack paradigm.
Specifically, we first introduce the experimental settings, which include the verifying tasks, datasets, metrics, models, and detailed experimental settings.

\subsection{Experimental Settings}

\subsubsection{Tasks and Datasets}
For supporting our conclusion about the feasibility of the proposed behavior backdoor attack, we choose multiple tasks as verification bench, following the principle of cross verification. Specifically, we select the image classification \cite{chen2021review}, the object detection \cite{zou2023object}, and the deepfake detection task \cite{yin2024improving}, respectively. As for the reasons, the image classification and object detection are the typical and widely-studies fundamental computer vision tasks, the deepfake detection is a classical safety-related scenario that worth investigation. Correspondingly, for each task, we select a representative dataset for performing experiments. More precisely, we take the image classification as the mainly-verifying task, thus 3 different but popular datasets are employed in the corresponding experiments, such as MNIST \cite{deng2012mnist}, CIFAR-10 \cite{krizhevsky2009learning}, and TinyImageNet \cite{chrabaszcz2017downsampled}. For object detection and deepfake detection tasks, we employ the PASCAL VOC 2007 \cite{everingham2010pascal} and Celeb-DF \cite{Li2019CelebDFAL}, respectively. We believe that the effectiveness of the proposed behavior backdoor could be well validated through this full consideration.

\subsubsection{Evaluation Metrics}
\label{sec:evaluationM}
Regarding the evaluation metrics for the proposed behavior backdoor attack, we define the attacking success rate (ASR) as basic assessment, which could be formulated as
\begin{equation*}
    \begin{aligned}
    \mathrm{ASR}=&\frac{\sum_{i\in \{i|{y}_i \neq \textit{target}\}}\mathcal{C}(x_i)}{N}, where \\
    \mathcal{C}(x_i)=&\left\{
    \begin{aligned}
        1, & if \quad\mathbb{F}_{\theta}(x_i) \neq y_{target} \land \mathbb{F}_{\theta^{*}}(x_i) = y_{target}, \\
        0, & others,
    \end{aligned}
    \right.
    \end{aligned}
\end{equation*}
where $\mathcal{C}(\cdot)$ is a counter that counts the sample size meeting the criteria, $N$ is the total size of the test set. This formula depicts that the ratio of successfully attacked samples of behavior backdoor to the total sample size. The higher the ASR, the better the attacking ability. In addition to the ASR that directly reflect attack capabilities, we also report the common metrics that correlated to the tasks themselves, such as accuracy (ACC) that widely used in classification tasks, mean Average Precision at IoU 0.5 (mAP@50) and F1 Score for detection tasks.

\subsubsection{Poisoned Models}
Verifying the effectiveness of our QB attack on different model architectures is of important significance in demonstrating the universality behavior backdoor. Therefore, for image classification and deepfake detection, we mainly consider 4 DLMs, \ie, AlexNet \cite{krizhevsky2012imagenet}, VGG \cite{simonyan2014very}, ResNet \cite{He2015DeepRL}, and ViT \cite{Dosovitskiy2020AnII}, that contains 2 typical architectures, \ie, convolutional neural networks (CNNs) and vision transformer models (ViTs). For object detection, we employ the FasterRCNN \cite{Ren2015FasterRT} and RetinaNet \cite{Lin2017FocalLF} for evaluation.

\subsubsection{Detailed Experimental Settings}

For training the poisoned backdoor model, we consistently employ Adaptive Moment Estimation optimizer (Adam)~\cite{Kingma2014AdamAM} with a  learning rate $1 \times 10^{-4}$ to minimize the loss function for updating the model parameters. Additionally, to ensure stable convergence of the model during training, we utilize the same learning rate decay approach as in CNNDetection~\cite{CNNDetection}. Specifically, the learning rate is reduced by a factor of 10 when the backdoor model's accuracy for the original labels fails to improve over several consecutive evaluations. The patience threshold for this decay is set to 7 by default. The mini-batch size $m$ is adjusted according to the specific dataset to facilitate effective training.  All experiments are conducted on a cluster equipped with NVIDIA GeForce RTX 3080Ti GPUs. The experiments are implemented in PyTorch version 2.3.0. 

\subsection{Attacking Performance}

In this section, we mainly report the attacking performance of the proposed QB attack on the image classification task.

Specifically, we conduct experiments on 3 different classical image classification related datasets. 4 selected models (AlexNet, VGG, ResNet, and ViT) are employed in these experiments. For each experiment, we report the ACC and ASR as mentioned in the Section \ref{sec:evaluationM}. Besides, we additionally provide the attack class accuracy (denoted as ``ACC$_{t}$''), which represents the accuracy of predicting an input sample as the target attacking class, \ie, $y_{target}$. Formally, the ACC$_{target} = \frac{\sum_{i}\mathbb{F}_{\theta^{*}}(x_i)=y_{target}}{N}$. The experimental results can be found in Table \ref{tab:attack-performance}, where we can draw a meaningful conclusion that the proposed QB attack, as a representative type of behavior backdoor, is feasible and effective and should be regarded as a fresh threat of reliable deep learning. We provide some insights as following:
\begin{itemize}
    \item The proposed idea of behavior backdoor, especially the quantification backdoor, is feasible for current DLMs. Taking the attacking results against ResNet on MNIST as an example, the ACC values of the vanilla model and the 3 poisoned model are 99.25\%, 99.27\%, 98.23\%, and 98.59\%, which are at a similar level. That means, before the poisoned models are triggered, they could successfully make accurate predictions, \ie, behaving good results. However, after triggering the poisoned backdoor models via the quantification method, we could witness significant differences. For instance, the ACC$_{t}$/ASR values under $attack_{target-0}$, $attack_{target-4}$, and $attack_{target-9}$ settings on VGG are respectively 100\%/99.21\%, 100\%/99.14\%, and 100\%/99.53\%, strongly supporting the effectiveness of our QB attack.
    \item The attacking ability of our QB attack shows differences on various datasets. Though the attacking performance is demonstrated consistently, we could find that the magnitudes of the attacking ability on MNIST, CIFAR-10, and TinyImageNet are quite different. For example, under the $attack_{target-0}$ setting, the ASR value on AlexNet of MNIST achieves 99.21\%, while that of  CIFAR-10 is 89.44\%, and that of TinyImageNet is 56.38\%. This clear decay tendency not only claims that the attacking ability of our behavior backdoor might be correlated to the dataset complexity, but also exhibits the large room for future improvement, calling for more efforts.
    \item The QB attack shows certain performance variations that are related to the model architectures. We could observe that the QB attack show distinguished ASR values on different models. Taking the ASR values on TinyImageNet as instances, the QB attack achieves 56.38\%, 68.67\%, 75.48\%, and 77.64\% on AlexNet, VGG, ResNet, and ViT, respectively. Upon this fact, we could find an interesting phenomenon that the better models appear lower defense against behavior backdoor, \ie, the ViT/AlexNet achieves 77.71\%/58.16\% ACC value on TinyImageNet under $attack_{target-0}$ setting, while the corresponding ASR values are 77.64\%/56.38\%. We conjecture that the stronger models might have more vulnerability against behavior backdoor due to their complex structures.
\end{itemize}

\begin{table*}[htb]
\caption{The attacking performance of our QB attack on different datasets and models. The ``vanilla'' indicates the benign models that are trained with normal settings, and the ``$attack_{target}$'' indicates the poisoned models that have been implanted backdoors, $-number$ is the index of the target class, \eg, $attack_{target-0}$ means the target attack class label index is 0.}
\label{tab:attack-performance}
\resizebox{\linewidth}{!}{
\begin{tabular}{cccccccccccccc}
\toprule
\multirow{2}{*}{Dataset}      & Model                    & \multicolumn{3}{c}{AlexNet} & \multicolumn{3}{c}{VGG} & \multicolumn{3}{c}{ResNet} & \multicolumn{3}{c}{ViT} \\ \cmidrule{2-14}
                              & Settings                 & ACC(\%)& ACC$_{t}$(\%)& ASR(\%)& ACC(\%)& ACC$_{t}$(\%)& ASR(\%)& ACC(\%)& ACC$_{t}$(\%)& ASR(\%)& ACC(\%)& ACC$_{t}$(\%)& ASR(\%)\\ \midrule
\multirow{4}{*}{MINIST}       & vanilla                  &       99.48&              -    & -    &     99.54&             -    & -   &      99.25&              -    & -    &     99.49&             -    & -   \\ \cmidrule{2-14}
                              & \textit{attack$_{target-0}$} &       99.26&              100.00&      99.21&     99.42&             100.00&     99.39&      99.27&              100.00&      99.21&     99.49&             100.00&     99.43\\\cmidrule{2-14}
                              & \textit{attack$_{target-4}$} &       99.34&              100.00&      99.35&     99.40&             100.00&     99.38&      98.23&              100.00&      98.14&     99.51&             100.00&     99.51\\\cmidrule{2-14}
                              & \textit{attack$_{target-9}$} &       98.94&              100.00&      98.99&     99.53&             100.00&     99.54&      98.59&              100.00&      98.53&     99.51&             100.00&     99.57\\ \midrule
\multirow{4}{*}{CIFAR-10}     & vanilla                  &       93.69&              -    & -    &     91.59&             -    & -   &      88.24&              -    & -    &     96.58&             -    & -   \\\cmidrule{2-14}
                              & \textit{attack$_{target-0}$} &       91.77&              97.95&      89.44&     91.55&             100.00&     91.39&      85.84&              100.00&      85.24&     96.74&             100.00&     96.63\\\cmidrule{2-14}
                              & \textit{attack$_{target-4}$} &       93.32&              100.00&      93.22&     91.34&             100.00&     91.39&      87.46&              100.00&      87.44&     96.66&             100.00&     96.60\\\cmidrule{2-14}
                              & \textit{attack$_{target-9}$} &       92.27&              99.88&      91.90&     90.31&             100.00&     89.84&      86.42&              99.99&      85.72&     96.43&             100.00&     96.47\\ \midrule
\multirow{4}{*}{TinyImageNet} & vanilla                  &       60.35&              -    &      -    &     70.88&             -    &     -    &      75.09&              -    &      -    &     76.71&             -    &     -    \\\cmidrule{2-14}
                              & \textit{attack$_{target-0}$} &       58.16&              97.75&      56.38&     68.76&             100.00&     68.67&      75.57&              100.00&      75.48&     77.71&             100.00&     77.64\\\cmidrule{2-14}
                              & \textit{attack$_{target-4}$} &       57.15&              96.90&      54.85&     68.98&             100.00&     69.06&      75.01&              100.00&      75.02&     77.55&             100.00&     77.57\\\cmidrule{2-14}
                              & \textit{attack$_{target-9}$} &       57.97&              98.38&      56.71&     68.34&             100.00&     68.36&      76.04&              100.00&      76.08&     77.28&             100.00&  77.31\\ \bottomrule  
\end{tabular}}
\end{table*}

\subsection{Ablation Study}

In our behavior backdoor model training process, there exists a hyperparameter, \ie, $\lambda$. It is worth investigating the effectiveness of the QB attack when facing different $\lambda$ to construct further understanding of this new-type QB attack paradigm. Thus, we conduct ablations by studying the attacking performance under multiple $\lambda$ values. 

\begin{figure}
     \centering
     \begin{subfigure}[MNIST]{0.49\linewidth}
         \centering
         \includegraphics[width=\linewidth]{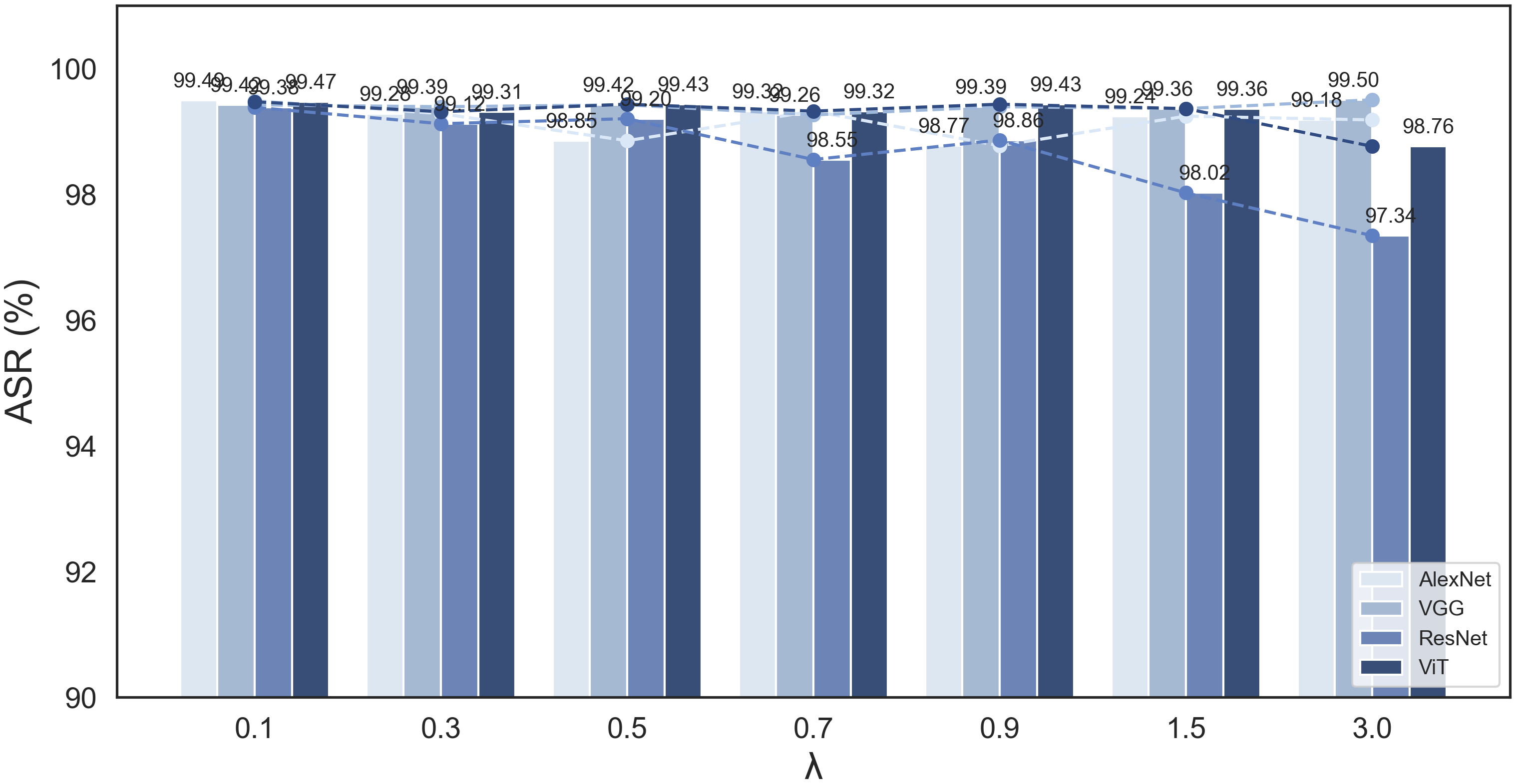}
     \end{subfigure}
    \begin{subfigure}[CIFAR-10]{0.49\linewidth}
         \centering
         \includegraphics[width=\linewidth]{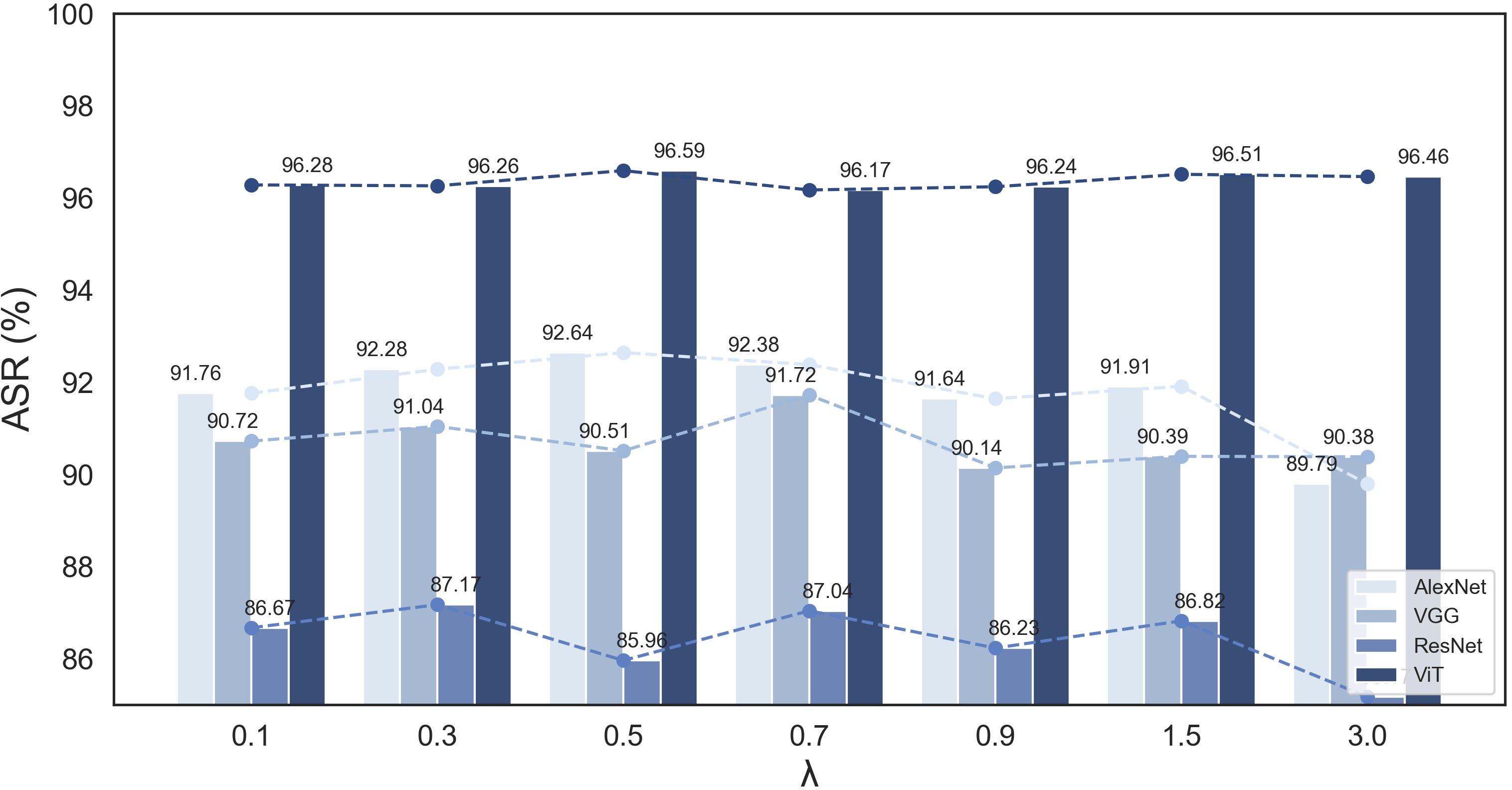}
     \end{subfigure}
     \caption{The ablation study on hyperparameter $\lambda$.}
     \label{fig:ablation}
 \end{figure}

Specifically, we set the $\lambda$ as 0.1, 0.3, 0.5, 0.7, 0.9, 1.5 and 3.0 respectively, to control the weight of the behavior backdoor related loss term $\mathcal{L}_{qba}$. The higher $\lambda$ value indicates the larger weight. We adopt the same settings of models and datasets with those of the attacking performance evaluation experiments. The experimental results can be witnessed in Figure \ref{fig:ablation}. It can be concluded that the value of $\lambda$ only make limited effects (\ie, only ResNet on MNIST, it appears more than 2\% ASR difference) on the attacking performance of QB attack, demonstrating the stability of the proposed method.

\subsection{Analysis and Discussion}
Beyond the aforementioned experiments, we are also interested in further answering 3 more research questions: \ding{182} how the proposed QB attack make effects? \ding{183} Could QB attack different tasks? \ding{184} If we adopt different quantification methods that are not employed during backdoor training, will they trigger the poisoned model? We then conduct additional experiments and give more analysis.

\subsubsection{Effectiveness of the Behavior Backdoor}
To investigate the effect mechanism of the proposed behavior backdoor attack, \ie, QB attack, we refer previous studies \cite{wang2022defensive} and conduct quantitative and qualitative analysis  by introducing the t-SNE \cite{van2008visualizing} and model attention \cite{selvaraju2017grad} tools, in which the former visualizes the feature extraction results of models and the latter analyzes the saliency regions that model focuses. 

\begin{figure}
     \centering
     \begin{subfigure}[t-SNE]{0.62\linewidth}
         \centering
         \includegraphics[width=\linewidth]{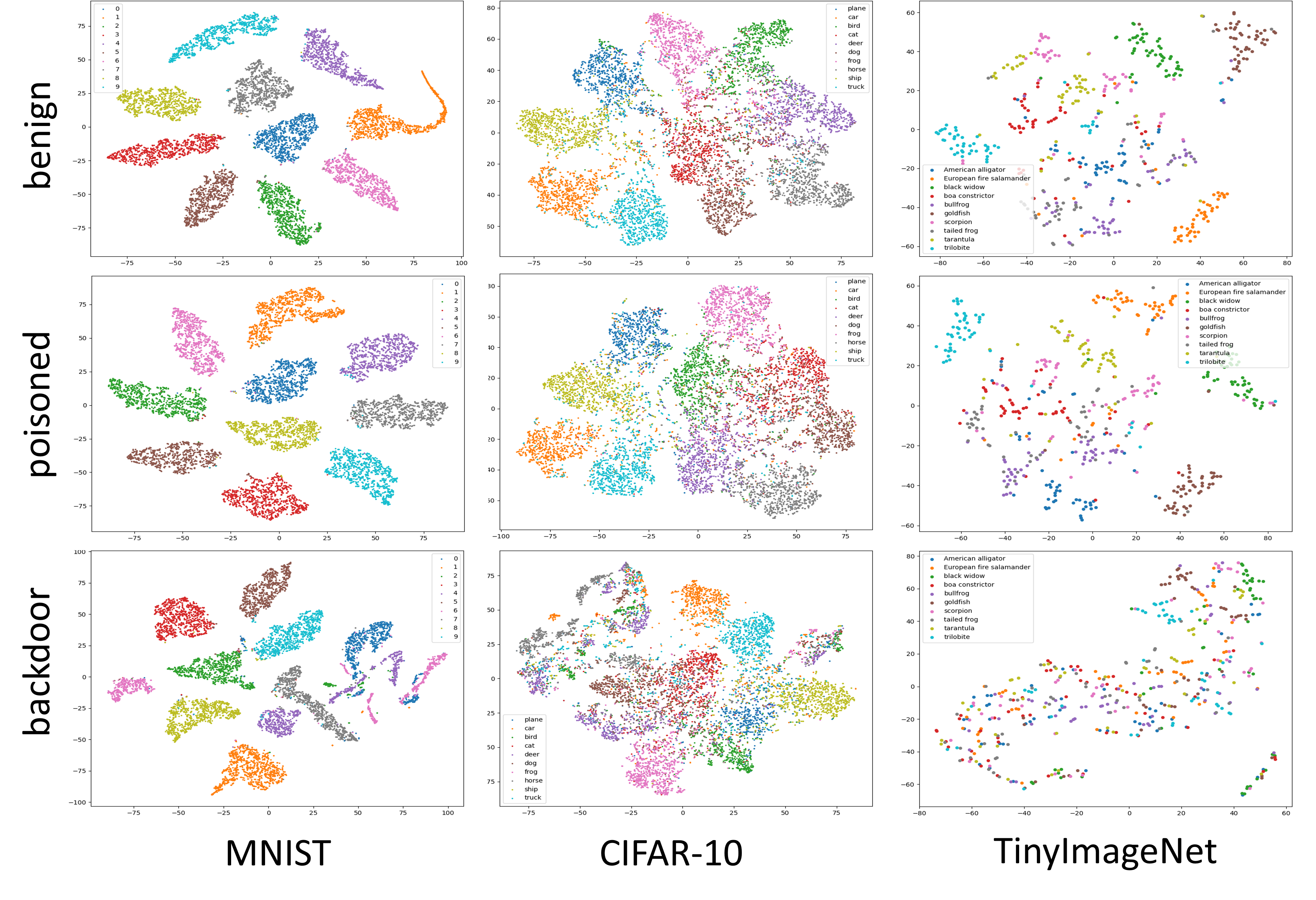}
     \end{subfigure}
    \begin{subfigure}[Attention]{0.35\linewidth}
         \centering
         \includegraphics[width=\linewidth]{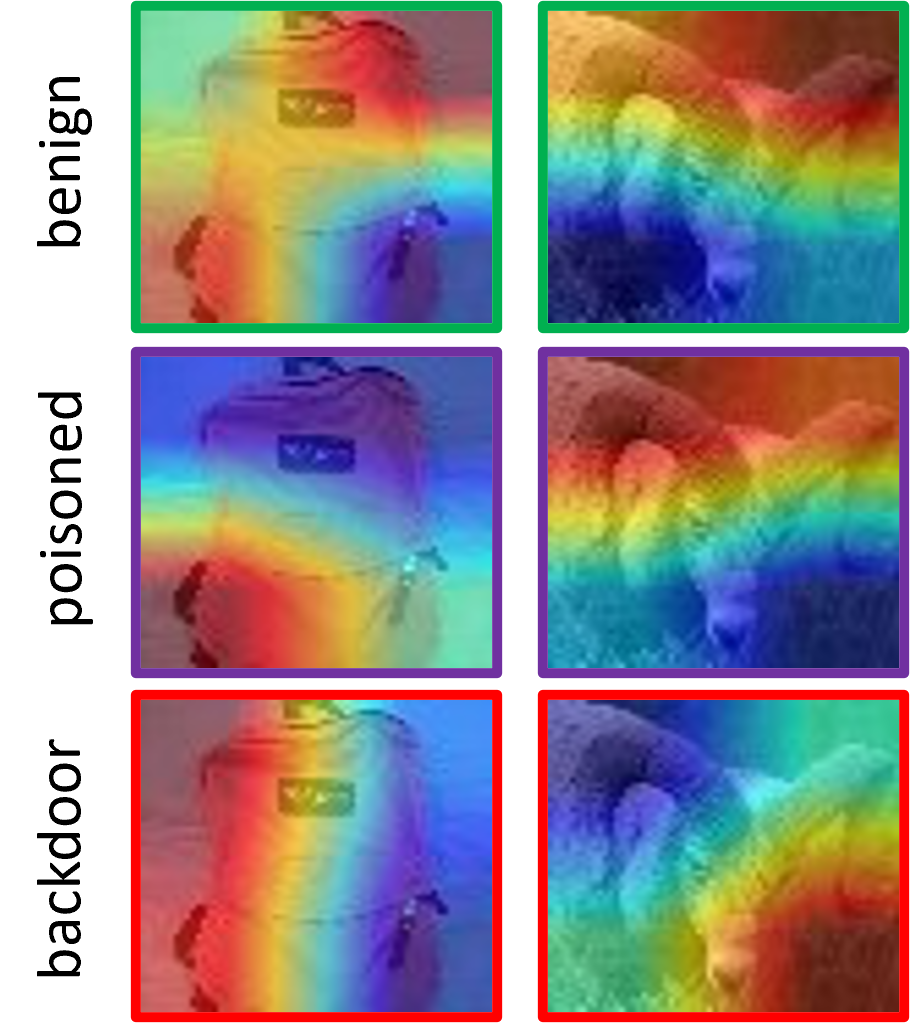}
     \end{subfigure}
     \caption{The t-SNE and model attention analysis. The ``benign'', ``poisoned'', and ``backdoor'' columns respectively  indicate the saliency map of benign, backdoor implanted but not triggered, and that of backdoor triggered model.}
     \label{fig:tsne-attention}
 \end{figure}

Specifically, \emph{for t-SNE analysis}, we plot the feature visualization of all benign models and poisoned models on MNIST, and CIFAR-10 datasets. The results can be found in Figure \ref{fig:tsne-attention}, left, where we find that the QB attack indeed change the learned feature distribution of the poisoned models. Taking the results on CIFAR-10 as instances, the feature distribution distance of target class and other class, \ie, green points, becomes closer. We thus attribute that the poisoned model make effects via adjusting the models' learning progress. 
For \emph{model attention analysis}, we employ all 4 models, randomly sample 10 instances from TinyImageNet, and then we conduct the saliency map generation according to the middle-layer feature maps following previous studies \cite{wang2021dual,wu2024napguard}. The results can be found in Figure \ref{fig:tsne-attention}, right, where we can conclude that the triggered backdoor models will pay no attention to the regions of key objects. More experimental results can be found in supplementary files.

\subsubsection{Attack Performance on Different Tasks}

Since our main experiments are conducted on the image classification task, it is also important to verify the attacking performance on different tasks. For this purpose, we conduct extra experiments on object detection task, and deepfake detection task.
Due to the space limitation, we provide the results on object detection task in supplementary files.

In detail, 
%
we adopt the CeleB-DF \cite{Li2019CelebDFAL} as benchmark and employ ResNet and ViT for evaluation. According to Table \ref{tab:deepfake}, it could be observed that before the backdoors in poisoned model are triggered, the ACC values of ResNet/ViT on deepfake detection task are 98.42\%/95.34\%, and after attack but not triggered, there ACC values are kept at similar level, while after the backdoors are triggered, the ASR values achieve high level, \ie, for ResNet, the ASR values under $attack_{target-real}$/$attack_{target-fake}$ are 98.47\%/97.35\%, demonstrating the attacking performance.
Thus, we could confirm that the QB attack, as a typical example of behavior backdoor attack, enjoys the feasibility and good attacking ability on various tasks.


\begin{table}[]
\centering
\caption{The attacking ability on deepfake detection.}
\label{tab:deepfake}
\resizebox{\columnwidth}{!}{%
\begin{tabular}{cccccc}
\toprule
\multirow{2}{*}{Metric} & \multirow{2}{*}{Model} & \multicolumn{3}{c}{Settings} \\ \cmidrule{3-5}
                        &                        & vanilla  & \textit{attack$_{target-real}$} & \textit{attack$_{target-fake}$} \\ \midrule
\multirow{2}{*}{ACC(\%)}    & ResNet                 & 98.42    & 92.76                        & 95.10                            \\ \cmidrule{2-5}
                        & ViT                    & 95.34    & 94.80                         & 93.00                            \\ \midrule
\multirow{2}{*}{ASR(\%)}    & ResNet                 & -        & 98.47                        & 83.51                           \\ \cmidrule{2-5}
                        & ViT                    & -        & 97.35                        & 74.63                           \\ \bottomrule
\end{tabular}%
}
\end{table}

\subsubsection{Triggering by Multiple Quantification Methods}

Since we first propose to exploit the post-training behavior, \ie, quantification, as the backdoor triggers, it is reasonable for us to ask if the quantification methods that are different from the behavior trigger $\mathcal{Q}$ during training could trigger the backdoors in the poisoned models. Thus, we try to trigger the backdoor by different quantification methods.

Specifically, we simply train a poisoned backdoor model with a set behavior trigger $\mathcal{Q}_{1}$ and then we utilize other quantification methods, namely DoreFa-Net \cite{zhou2016dorefa1} (denoted as ``$\mathcal{Q}_{2}$'') and Low-bit Quantization \cite{li2016ternary} (denoted as ``$\mathcal{Q}_{3}$''), to test if the backdoors inside the trained model will be activated. The corresponding experiments are conducted on CIFAR-10. The results can be found in Table \ref{tab:quant}, where we can draw some meaningful conclusions: \ding{182} under white-box settings, \ie, the behavior trigger $\mathcal{Q}$ in testing is same with that in training, the behavior backdoor performs well as we mentioned before. More precisely, the ASR values of the diagonal columns in the table are 85.24\%, 86.99\%, 86.10\%, from top left to bottom right. \ding{183} Under black-box settings, \ie, $\mathcal{Q}$ in testing is different from that in training, the ASR values appear clear drop, \ie, when using $\mathcal{Q}_1$/$\mathcal{Q}_2$ to trigger the $\mathcal{Q}_{3}$ trained backdoor models, the ASR only achieves 0.09\%/0.10\%. This fact indicates the weak adaptation of our behavior backdoor to different quantification triggers. However, it could be also observed that in some cases, the cross-quantification triggers achieve a relatively high AS values. For instance, $\mathcal{Q}_{3}$ could trigger the $\mathcal{Q}_{1}$/$\mathcal{Q}_{2}$ trained backdoor models and achieves 83.60\%/45.30\% ASR, which reveals that the transferable behavior backdoor could be a future research direction. 

\begin{table}[]
\caption{The results on triggering the backdoor with different quantification methods. The experiments are conducted on CIFAR-10.}
\label{tab:quant}
\resizebox{\linewidth}{!}{
\begin{tabular}{ccccccc}
\toprule
Quantification & \multicolumn{2}{c}{$\mathcal{Q}_{1}$} & \multicolumn{2}{c}{$\mathcal{Q}_{2}$} & \multicolumn{2}{c}{$\mathcal{Q}_3$}\\ \cmidrule{2-7}
Method         & ACC$_{t}$(\%)& ASR(\%)& ACC$_{t}$(\%)& ASR(\%)& ACC$_{t}$(\%)& ASR(\%)\\ \midrule
$\mathcal{Q}_{1}$            &100.00&85.24&64.71     &51.30&97.69    &83.60\\ \midrule
$\mathcal{Q}_{2}$  &10.09     &0.09     &100.00&86.99  &52.67    &45.30\\ \midrule
$\mathcal{Q}_{3}$            & 10.23    &0.07     & 10.19    &0.10&100.00&86.10\\ \bottomrule
\end{tabular}}
\end{table}

\section{Conclusion}
\label{sec:con}

In this paper, we take the first step towards behavior backdoor attack paradigm and propose the first practical pipeline of exploiting quantification operation as backdoor trigger, \ie, the quantification backdoor (QB) attack, by elaborating the behavior-driven backdoor object optimizing and address-shared backdoor model training method. Extensive experiments on multiple dataset with different models are conducted for various tasks, strong demonstrating the feasibility of the proposed concept of behavior backdoor attack and the effectiveness of the QB attack. 

\textbf{Ethical statements}. Since a new-type attacking method against DLMs is proposed, 
it is necessary to concern the possible influence to the deep learning communities. Thus, 
we state the potential impacts as: \ding{182} behavior backdoor for social bad. As an attacking approach, the QB attack might be exploited by hackers to manuscript poisoned models and then to drop them into public fields for their ulterior goals, like destroying the edge device deployment or inducing specific results. \ding{183} Behavior backdoor for social good. Though harmful, it could be also utilized in bona fide intentions, such as protect the self-developed models from unauthorized use, to achieve the goal of neither affecting open source nor losing profits. \ding{184} Responsible disclosure. Considering the potential ethical risks, we only support this study to be exploited for research goals. Therefore, we will provide the codes, datasets, and checkpoints, for the scholars after communicating and reviewing.

\clearpage
{
    \small
    \bibliographystyle{ieeenat_fullname}
    \bibliography{main}
}


\end{document}


\maketitle

In our main paper, due to the space limitation, we could not provide more extensive evidence to readers for supporting the demonstration of the proposed \textbf{behavior backdoor}, especially the \emph{quantification backdoor (\textbf{QB}) attack}. In this file, we aim to provide more experiments to prove it. The rest of this file includes the detailed algorithm, additional results on objection task, more ablations study results, and extra analysis results of the proposed QB attack.

\section{The overall training process}
As we mentioned in the main paper, for a certain dataset $\mathcal{X}$, we first select a quantification method and take it as the behavior backdoor trigger $\mathcal{Q}$. Then we calculate the loss function $\mathcal{L}_{qba}$ value for each input $x_i$, and we optimize the $\mathbb{F}_{\theta}$ following an updated objective function $\arg\min_{\theta^{*}}\mathcal{L}_{overall}$ Note that the $\theta^{*}$ is aligned to the $\theta$, thus $\mathbb{F}_{\theta}$ could be trained as expected. The detailed algorithm is shown in Algorithm \ref{alg:algorithm}.

\begin{algorithm}[h]
    \caption{QBA algorithm}
    \label{alg:algorithm}

    	\KwIn{backdoor model $\mathbb{F}_{\theta}$, quantification operator $\mathcal{Q}, $dataset $\mathcal{X}$, a weighting factors $\lambda$, maximum iteration number $T$, mini-batch size $m$.}
    	\KwOut{Trained model $\mathbb{F}_{\theta}$.}
        $\mathbb{F}_{\theta^{*}}\gets\mathcal{Q}(\mathbb{F}_{\theta})$; \\
        $\mathcal{P}_{\theta}\gets\mathcal{S}(\mathbb{F}_{\theta})$ \Comment*[r]{Parameter list}
        $\mathcal{P}_{\theta^{*}}\gets\mathcal{S}(\mathbb{F}_{\theta^{*}})$ \Comment*[r]{Parameter list}
        $\mathcal{A}(\mathcal{P}_{\theta}, \mathcal{P}_{\theta^{*}})$ \Comment*[r]{Sharing address}
        \While{$t < T$}
        {
            $x_{t} \sim \mathcal{X}: |x_{t}|=m$ \Comment*[r]{Randomly sample}
            $g \gets \mathop{\mathds{E}}\limits_{x \sim x_{t}}[\nabla_{\theta^{*}}\mathcal{L}_{overall}]$  \Comment*[r]{Gradient}
            $\theta^{*} \gets \mathrm{Optim}(g$)$ $\Comment*[r]{Update}
        }
\end{algorithm}

\subsection{More experiments on object detection task}

In our main paper, we report the experimental result on image classification and deepfake detection. Here, we conduct additional experiments on object detection task. 

Specifically, for the poisoned model training, we combine the classification loss and bounding box regression loss from each part to obtain the respective loss functions, $\mathcal{L}_{ben}$ and $\mathcal{L}_{qba}$. And we employ some typical models, \ie, the FasterRCNN and RetinaNet.
In this experiment, we set the weight parameter to 1, the overall loss is given by $\mathcal{L}_{overall} = \mathcal{L}_{ben} + \mathcal{L}_{qba}$. We adopt 2 different attacking target settings, where $attack_{target-car}$ means that the class ``\textit{car}'' is set as attacking target, $attack_{target-cat}$ means that the class ``\textit{cat}'' is set as attacking target. We adopt the widely-used mAP@50, and ASR as evaluation indicators. For ASR evaluation, we begin by filtering samples based on their IoU values, classifying those with IoU greater than 0.5 as correctly predicted and the rest as incorrectly predicted. Subsequently, ASR is calculated following the evaluation criteria outlined in 4.1.2 section.
The experimental  results can be found in Table \ref{tab:object}. Consistently, the mAP@50/ASR values of FasterRCNN/RetinaNet on object detection task appear the same phenomena with that in our main paper, \ie, the QB attack could satisfy the expectation of our design. More precisely, for vanilla model (\ie, the no-poisoned model), the mAP@50 achieves 40.71\%, while those of the poisoned models (\ie, $attack_{target-car}$ and $attack_{target-cat}$) are respectively 43.41\% and 41.38\%, maintaining at similar level.  Similarly, the ASR on $attack_{target-car}$\/$attack_{target-cat}$ are respectively 58.94\%/58.58\%, demonstrating the effectiveness of the QB attack. The similar results can be witnessed on the results of RetinaNet. 

\begin{table}[]
\caption{The attacking ability on object detection.}
\label{tab:object}
\resizebox{\linewidth}{!}{
\begin{tabular}{cccc}
\toprule
Model                & Settings                 & MAP@50(\%)& ASR(\%) \\ \midrule
\multirow{3}{*}{FasterRcnn} 
& vanilla                 & 40.71        &  -   \\ \cmidrule{2-4}
& \textit{attack$_{target-car}$} & 43.41        & 58.94    \\ \cmidrule{2-4}
& \textit{attack$_{target-cat}$} & 41.38        & 58.58    \\ \midrule
\multirow{3}{*}{RetinaNet} 
& vanilla                 & 38.90      &  -   \\ \cmidrule{2-4}
& \textit{attack$_{target-car}$} & 38.57       & 44.29    \\ \cmidrule{2-4}
& \textit{attack$_{target-cat}$} & 36.15        & 44.88   \\ \bottomrule
\end{tabular}}
\end{table}

\subsection{More experiments on ablation studies}

In this section, we provide additional ablation studies on a more popular dataset, \ie, TinyImageNet, to investigate the effectiveness of $\lambda$.

\begin{figure}[htbp]
\centering
\includegraphics[width=\linewidth]{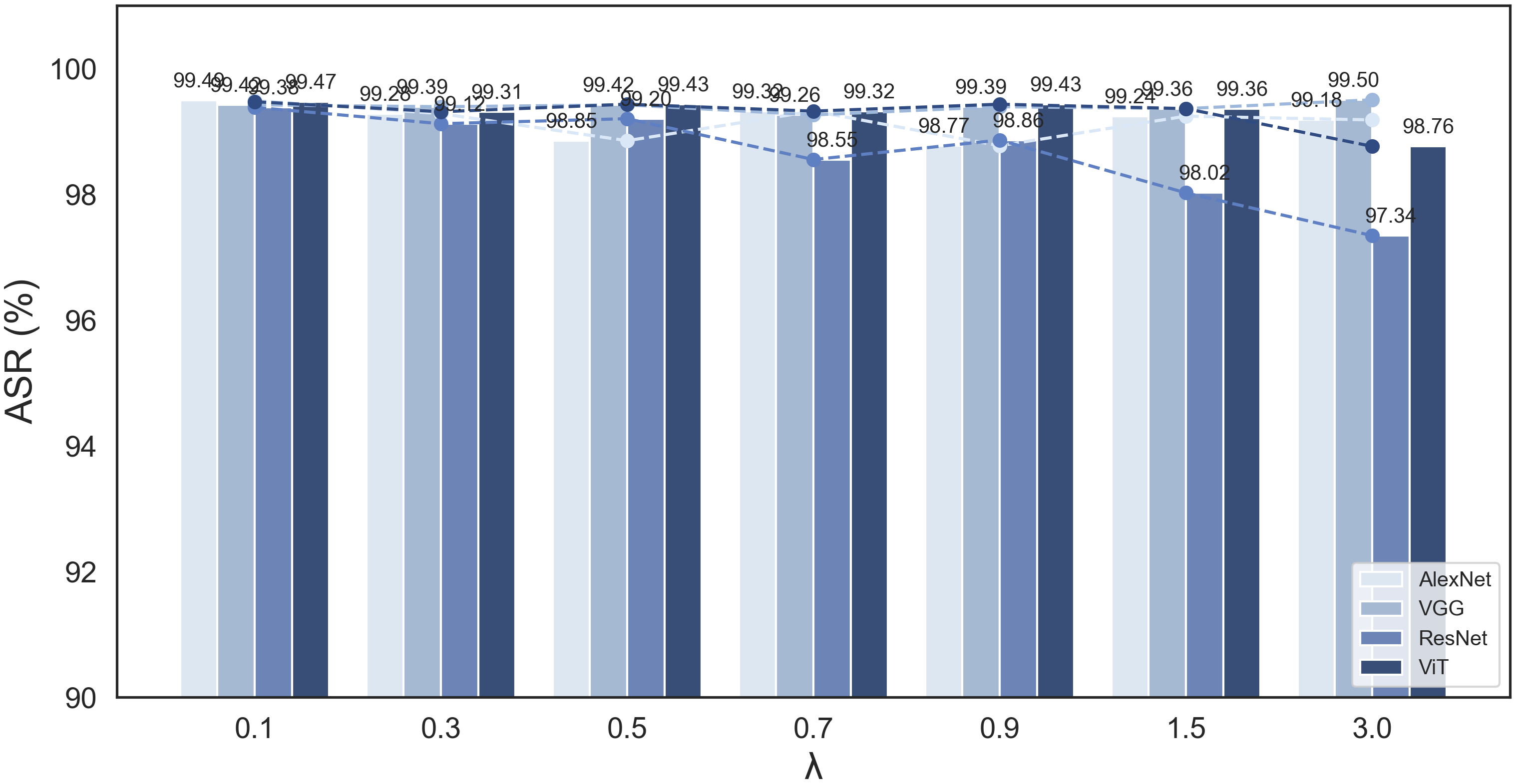}
\caption{The ablation of hyperparameter $\lambda$ on MNIST.}
\label{fig:MNIST}
\end{figure}

\begin{figure}[htbp]
\centering
\includegraphics[width=\linewidth]{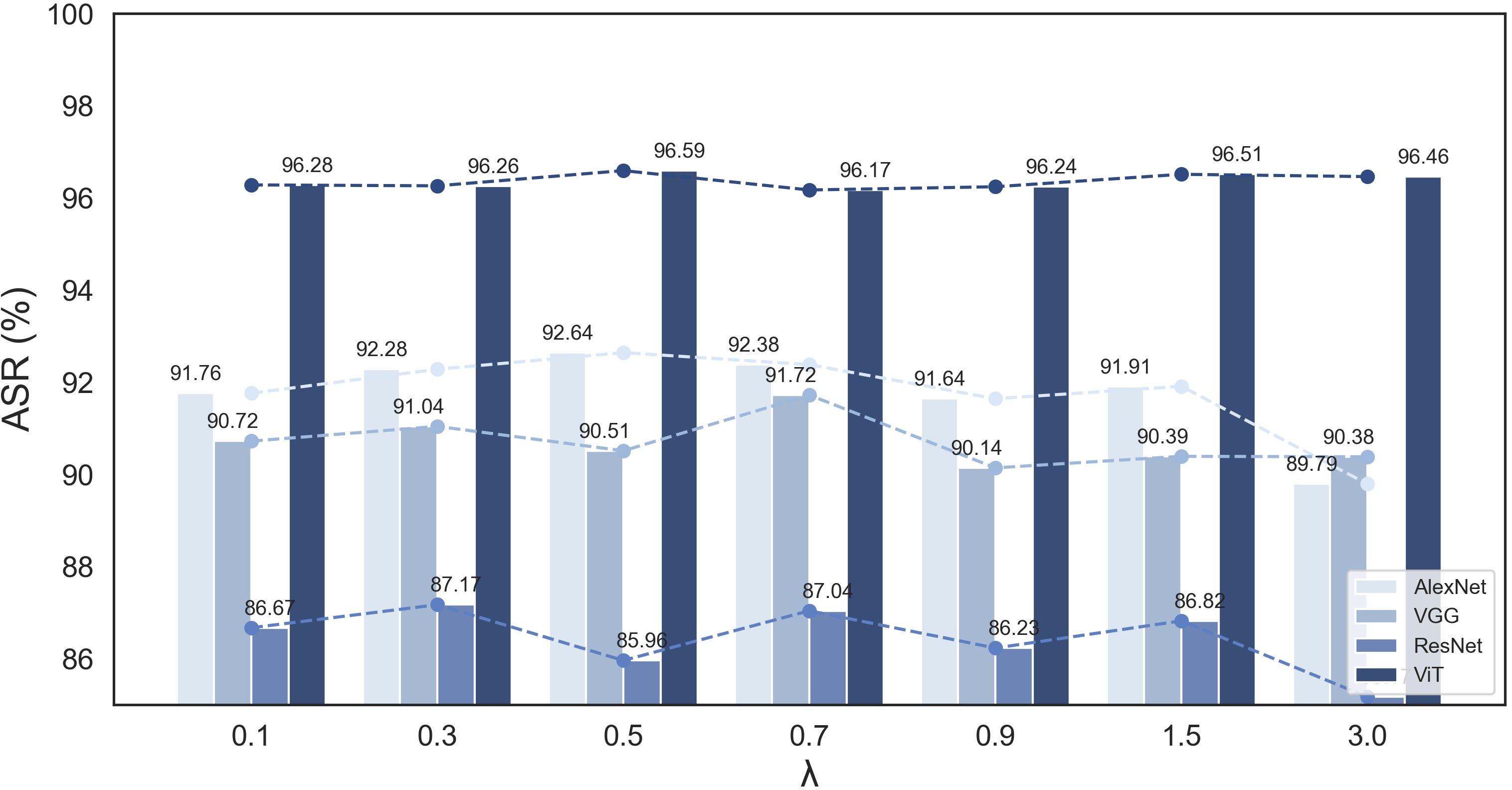}
\caption{The ablation of hyperparameter $\lambda$ on CIFAR-10.}
\label{fig:CIFAR}
\end{figure}

\begin{figure}[htbp]
\centering
\includegraphics[width=\linewidth]{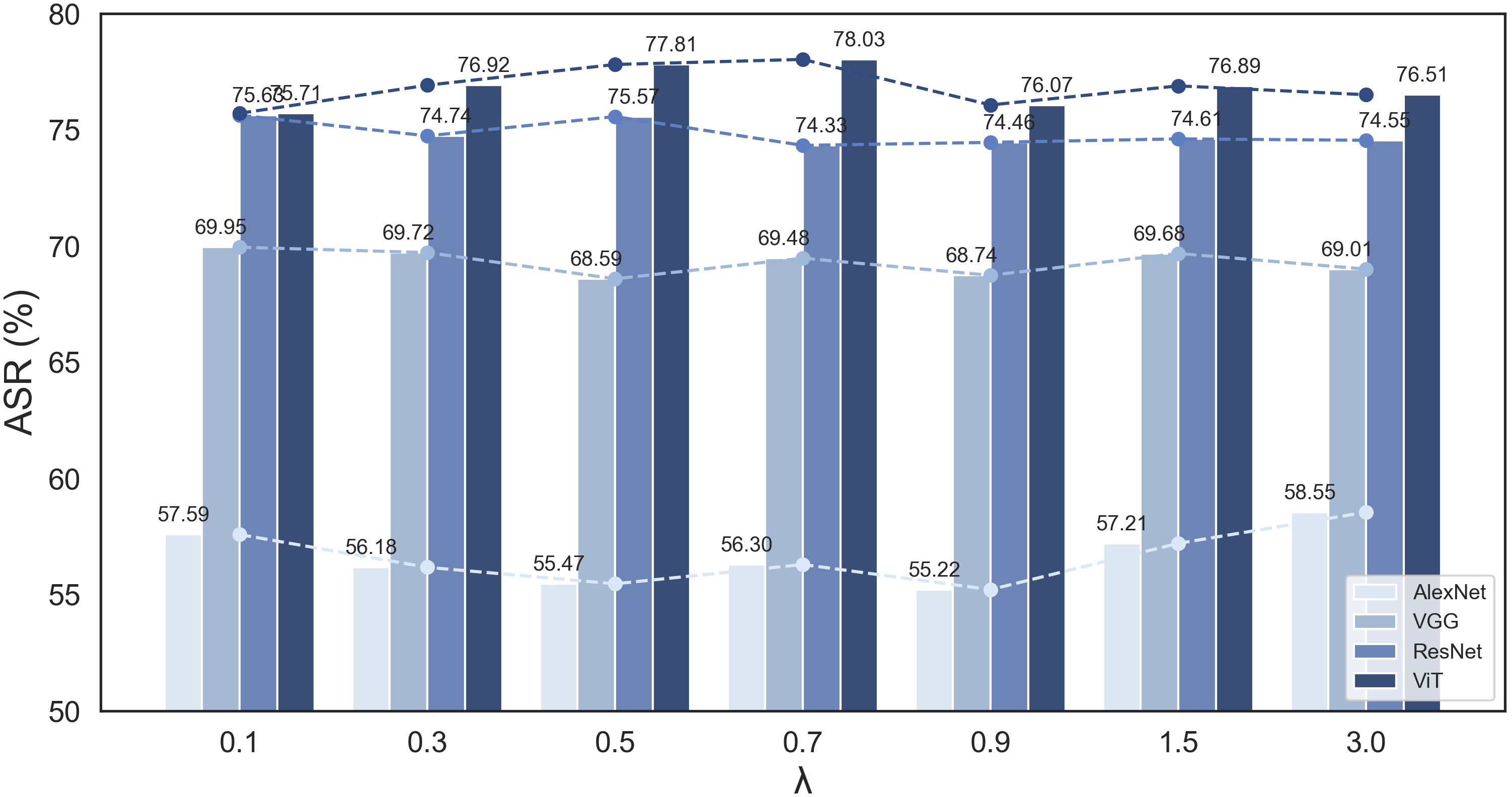}
\caption{The ablation of hyperparameter $\lambda$ on TinyImagenet.}
\label{fig:TinyImagenet}
\end{figure}

Specifically, we adopt similar experimental settings with that in the main paper, and the results are shown in Figure \ref{fig:MNIST}, Figure \ref{fig:CIFAR}, and Figure \ref{fig:TinyImagenet}.
Besides, to better observe the specific attacking values, we post the detailed results in Table \ref{tab:all-ablation}.
The corresponding results on various datasets, including MNIST, CIFAR-10, and TinyImageNet, strongly support the conclusion that $\lambda$ make limited effects only and demonstrates the stability of the proposed QB attack, revealing the great future potential of this never-discovered attacking mode. More precisely, {we have already discussed in Section 4.3 that only ResNet on MNIST exhibits an ASR difference exceeding 2\%. However, on the TinyImageNet dataset, the ASR difference is slightly larger, particularly for AlexNet, which achieves 3.33\%. We infer that this is related to the complexity of the dataset, given that TinyImageNet has 20 times the number of classes compared to the MNIST and CIFAR-10 dataset. Despite these minor increases, it still indicates that the influence of $\lambda$ on the model's ASR is quite limited. Furthermore, for ResNet and ViT, we find that smaller values of $\lambda$ tend to lead to a little better model performance. For the AlexNet and VGG models, this observation could be not witnessed, \ie, the best $\lambda$ might occur as high or low values}.



\begin{table}[]
\caption{Detailed results of the ablation studies. We only report the ASR values under different settings (\%). Additionally, the attacking ability tendency corresponding to the dataset scale, \ie , more complex and larger datasets pose strong challenges to the proposed behavior backdoor attack such as QB attack, has been once again demonstrated as we stated in the main paper.}
\label{tab:all-ablation}
\resizebox{\linewidth}{!}{
\begin{tabular}{lcccc}
\bottomrule
\multicolumn{2}{l}{\diagbox{Model}{Dataset}}    & MNIST & CIFAR-10 & \multicolumn{1}{c}{TinyImageNet} \\ \midrule
\multirow{7}{*}{AlexNet} & 0.1 &       99.49&          91.76&                                  57.59\\ \cmidrule{2-5}
                         & 0.3 &       99.28&          92.28&                                  56.18\\ \cmidrule{2-5}
                         & 0.5 &       98.85&          92.64&                                  55.47\\ \cmidrule{2-5}
                         & 0.7 &       99.32&          92.38&                                  56.30\\ \cmidrule{2-5}
                         & 0.9 &       98.77&          91.64&                                  55.22\\ \cmidrule{2-5}
                         & 1.5 &       99.24&          91.91&                                  57.21\\ \cmidrule{2-5}
                         & 3   &       99.18&          89.79&                                  58.55\\ \midrule
\multirow{7}{*}{VGG}     & 0.1 &       99.42&          90.72&                                  69.95\\ \cmidrule{2-5}
                         & 0.3 &       99.39&          91.04&                                  69.72\\ \cmidrule{2-5}
                         & 0.5 &       99.42&          90.51&                                  68.59\\ \cmidrule{2-5}
                         & 0.7 &       99.26&          91.72&                                  69.48\\ \cmidrule{2-5}
                         & 0.9 &       99.39&          90.14&                                  68.74\\ \cmidrule{2-5}
                         & 1.5 &       99.36&          90.39&                                  69.68\\ \cmidrule{2-5}
                         & 3   &       99.50&          90.38&                                  69.01\\ \midrule
\multirow{7}{*}{ResNet}  & 0.1 &       99.38&          86.67&                                  75.63\\ \cmidrule{2-5}
                         & 0.3 &       99.12&          87.17&                                  74.74\\ \cmidrule{2-5}
                         & 0.5 &       99.20&          85.96&                                  75.57\\ \cmidrule{2-5}
                         & 0.7 &       98.55&          87.04&                                  74.33\\ \cmidrule{2-5}
                         & 0.9 &       98.86&          86.23&                                  74.46\\ \cmidrule{2-5}
                         & 1.5 &       98.02&          86.82&                                  74.61\\ \cmidrule{2-5}
                         & 3   &       97.34&          85.17&                                  74.55\\ \midrule
\multirow{7}{*}{ViT}     & 0.1 &       99.47&          96.28&                                  75.71\\ \cmidrule{2-5}
                         & 0.3 &       99.31&          96.26&                                  76.92\\ \cmidrule{2-5}
                         & 0.5 &       99.43&          96.59&                                  77.81\\ \cmidrule{2-5}
                         & 0.7 &       99.32&          96.17&                                  78.03\\ \cmidrule{2-5}
                         & 0.9 &       99.43&          96.24&                                  76.07\\ \cmidrule{2-5}
                         & 1.5 &       99.36&          96.51&                                  76.89\\ \cmidrule{2-5}
                         & 3   &       98.76&          96.46&            76.51\\ \bottomrule                   
\end{tabular}}
\end{table}

\section{Extra analysis results of the QB attack}
In this section, we provide extra results of the QB attack regarding the t-SNE analysis and model attention analysis.

\begin{figure}[htbp]
\centering
\includegraphics[width=\linewidth]{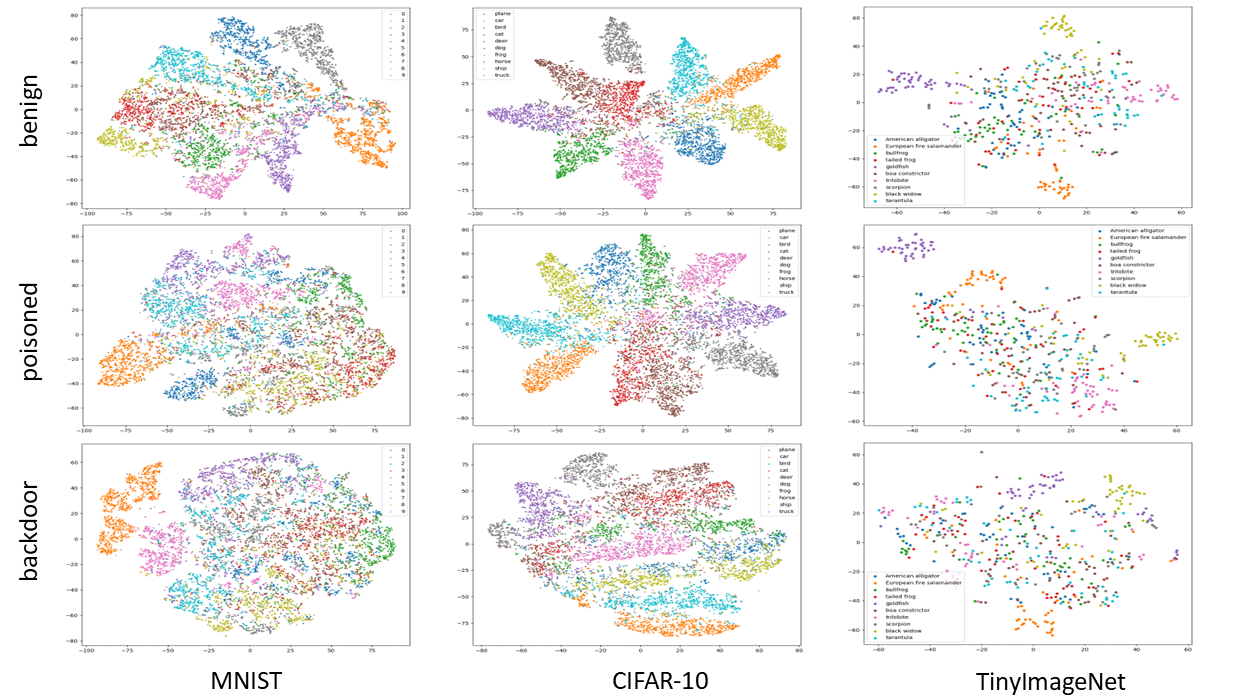}
\caption{t-SNE analysis on AlexNet model}
\label{fig:alex}
\end{figure}

\begin{figure}[htbp]
\centering
\includegraphics[width=\linewidth]{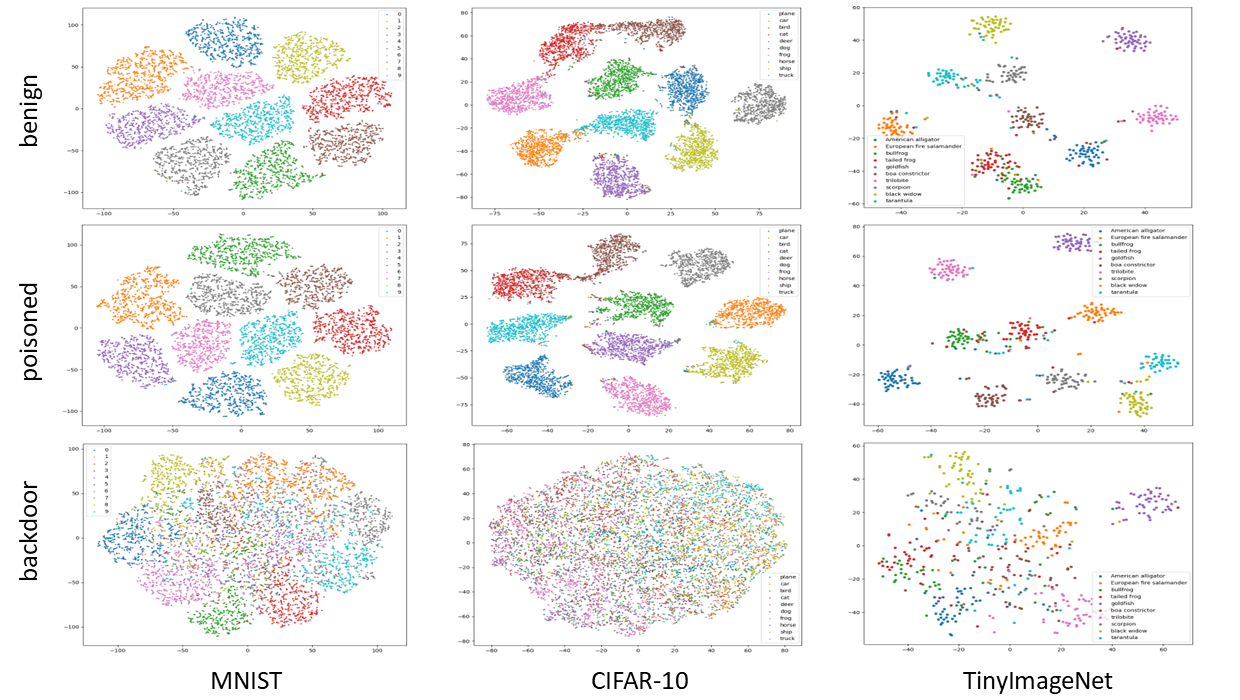}
\caption{t-SNE analysis on ViT model}
\label{fig:vit}
\end{figure}

\begin{figure}[htbp]
\centering
\includegraphics[width=\linewidth]{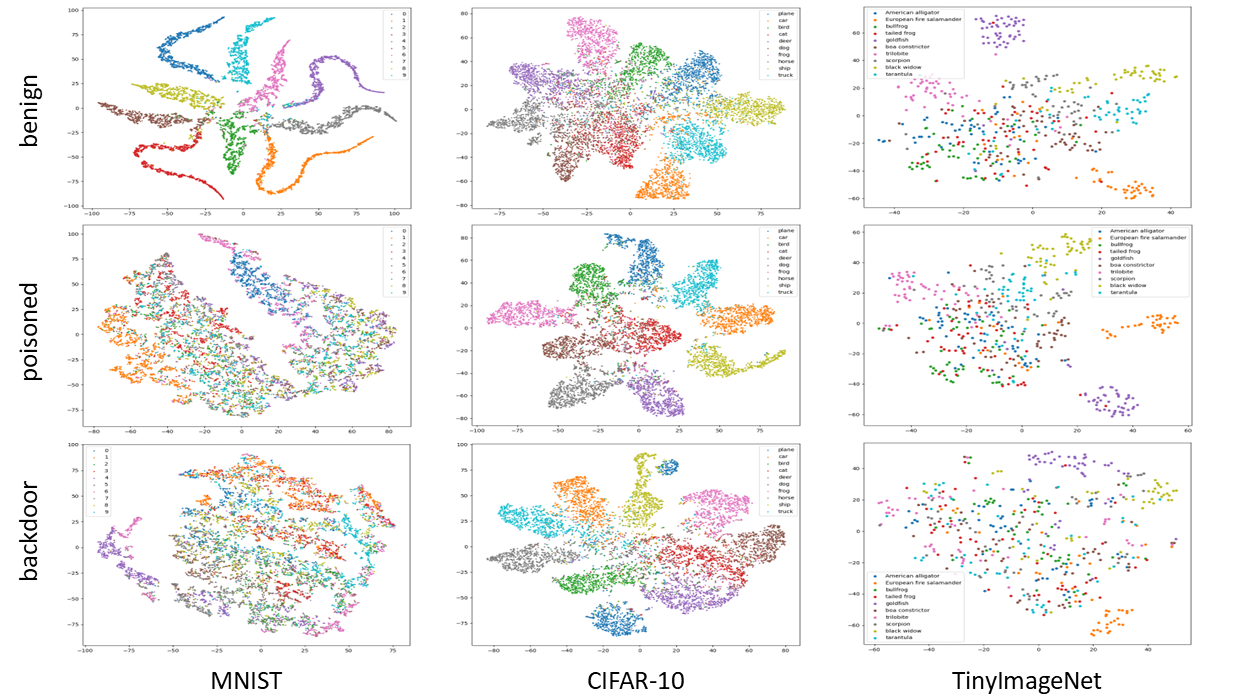}
\caption{t-SNE analysis on VGG model}
\label{fig:vgg}
\end{figure}

Specifically, for the t-SNE analysis, we only report the corresponding results on ResNet in the regular manuscript. Hence, we here provide more results on AlexNet, ViT, and VGG models as shown in Figure \ref{fig:alex}, Figure \ref{fig:vit}, and Figure \ref{fig:vgg}. Note that we still adopt the image classification task and employ all 3 corresponding datasets in our main paper. As a result, a consistent phenomenon could be witnessed on different datasets, namely, before and after backdoor implanting, the model learned feature distribution has been significantly influenced, which appears in feature distance between different class. And before and after backdoor triggering, the model extracted feature has been greatly confused, \ie, the entire distribution of all classes are totally mixture, resulting in false predictions. To be specific, {take the results on CIFAR-10 as examples, in benign models, data points of the target class (\ie, the deep blue points) are usually well separated from other categories, and the decision distance is very clear, which can predict good results. In the poisoned model, due to the insertion of a malicious trigger, the data point distribution of the target class becomes blurred, reducing the separation from other classes, but still showing overall separability. In the backdoor model, after the backdoor is triggered, the target class appears fuzzy, for instance, there occurs two deep blue point groups, indicating that the feature clustering is totally disturbed to result in wrong predictions}.

\begin{figure}[htbp]
\centering
\includegraphics[scale=0.57]{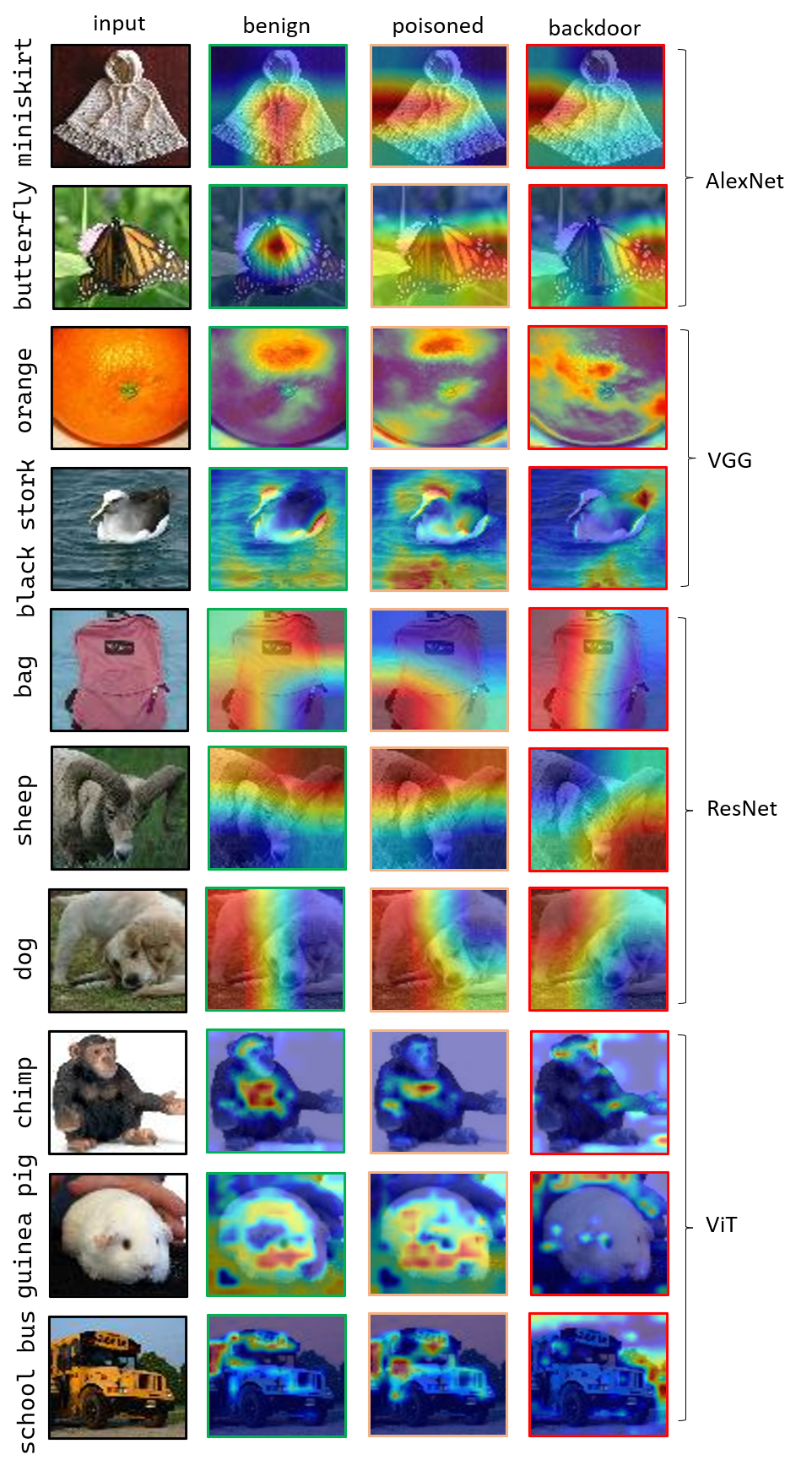}
\caption{Model attention analysis. Specifically, the first two rows focus on the analysis of the AlexNet model, the third and fourth rows focus on the VGG model, the fifth to seventh rows focus on the ResNet model, and the final three rows focus on the ViT model.}
\label{fig:attention-supp}
\label{figure}
\end{figure}

Regarding the model attention analysis, we conduct additional experiments considering the model variety and sample variety. Specifically, we only provide 2 attention analysis instances in the main paper, it is necessary for us to introduce more instances and model architectures to give stronger evidence for helping to understand the behavior mechanism of the QB attack. Specifically, for class \texttt{miniskirt} and \texttt{monarch butterfly} (denoted as ``\texttt{butterfly}''), we conduct attention analysis on AlexNet model. For class \texttt{orange} and \texttt{black stork}, we conduct attention analysis on VGG model. For class \texttt{bad}, \texttt{sheep}, and \texttt{dog}, we conduct attention analysis on ResNet model. For class \texttt{chimp}, \texttt{guinea pig}, and \texttt{school bus}, we conduct attention analysis on ViT model. The visualization can be found in Figure \ref{fig:attention-supp}, where we can draw the same conclusion, \ie, the triggered behavior backdoors that are generated by QB attack affect the model global perception of the instances, with that in our main paper.

In detail, taking \texttt{guinea pig} category as an instance, we can witness the attention region variance in different status. In benign and poisoned models (with backdoor but not been triggered) is primarily concentrated on the body of the input \texttt{guinea pig} sample, while their attention regions have certain distribution differences.  However, the attention of backdoor model (with backdoor been triggered) shifts significantly toward the surrounding environment, which might result in wrong prediction accordance. Further, it can be found that the backdoor performance of ResNet and ViT models notably surpasses that of AlexNet and VGG, \ie, the ``hot'' regions are almost totally been transferred into the corner regions. This phenomenon aligns with our observation in Section 4.2 that models with more complex architectures and superior performance are more vulnerable to behavioral backdoor attacks. We conjecture the reason behind this phenomenon is the more powerful models learn more accurate but trigger-sensitive knowledge during training, thus activating more discrete attention region distributions as a result.

\clearpage
